\pdfoutput=1

\documentclass[11pt]{article}

\usepackage{emnlp2021}

\usepackage{times}
\usepackage{latexsym}

\usepackage[T1]{fontenc}

\usepackage[utf8]{inputenc}

\usepackage{microtype}
\usepackage{graphicx}
\usepackage{amsmath}
\usepackage{amssymb}
\usepackage{amsthm}
\usepackage{booktabs}
\usepackage{algorithm}
\usepackage{algorithmic}
\usepackage{multirow}
\usepackage{bm}

\newcommand\blfootnote[1]{%
  \begingroup
  \renewcommand\thefootnote{}\footnote{#1}%
  \addtocounter{footnote}{-1}%
  \endgroup
}

\title{Progressively Guide to Attend: An Iterative Alignment Framework for Temporal Sentence Grounding}

\author{Daizong Liu\textsuperscript{1,2}, Xiaoye Qu\textsuperscript{3}, Pan Zhou\textsuperscript{1*} \\
  \textsuperscript{1}The Hubei Engineering Research
Center on Big Data Security, School of \\ Cyber Science and Engineering, Huazhong University of Science and Technology \\
  \textsuperscript{2}School of Electronic Information and Communication, \\ Huazhong University of Science and Technology \\
  \textsuperscript{3}Huawei Cloud \\
  {\tt \normalsize dzliu@hust.edu.cn, quxiaoye@huawei.com, panzhou@hust.edu.cn}}

\begin{document}
\maketitle
\begin{abstract}
A key solution to temporal sentence grounding (TSG) exists in how to learn effective alignment between vision and language features extracted from an untrimmed video and a sentence description.
Existing methods mainly leverage vanilla soft attention to perform the alignment in a single-step process.
However, such single-step attention is insufficient in practice, since complicated relations between inter- and intra-modality are usually obtained through multi-step reasoning. 
In this paper, we propose an Iterative Alignment Network (IA-Net) for TSG task, which iteratively interacts inter- and intra-modal features within multiple steps for more accurate grounding.
Specifically, during the iterative reasoning process, we pad multi-modal features with learnable parameters to alleviate the nowhere-to-attend problem of non-matched frame-word pairs, and enhance the basic co-attention mechanism in a parallel manner. To further calibrate the misaligned attention caused by each reasoning step, we also devise a calibration module following each attention module to refine the alignment knowledge.
With such iterative alignment scheme,
our IA-Net can robustly capture the fine-grained relations between vision and language domains step-by-step for progressively reasoning the temporal boundaries. Extensive experiments conducted on three challenging benchmarks demonstrate that our proposed model performs better than the state-of-the-arts.
\vspace{-10pt}
\end{abstract}
\blfootnote{
\textsuperscript{*}Corresponding author.}

\section{Introduction}
Temporal localization is an important topic of visual understanding in computer vision. There are several related tasks proposed for different scenarios involving language, such as video summarization \cite{song2015tvsum,chu2015video}, temporal action localization \cite{shou2016temporal,zhao2017temporal}, and temporal sentence grounding \cite{gao2017tall,anne2017localizing}. Among them, temporal sentence grounding is the most challenging task due to its complexity of multi-modal interactions and complicated context information. Given an untrimmed video, it aims to determine the segment boundaries including start and end timestamps that contain the interested activity according to a given sentence description.

\begin{figure}[t!]
\centering
\includegraphics[width=0.48\textwidth]{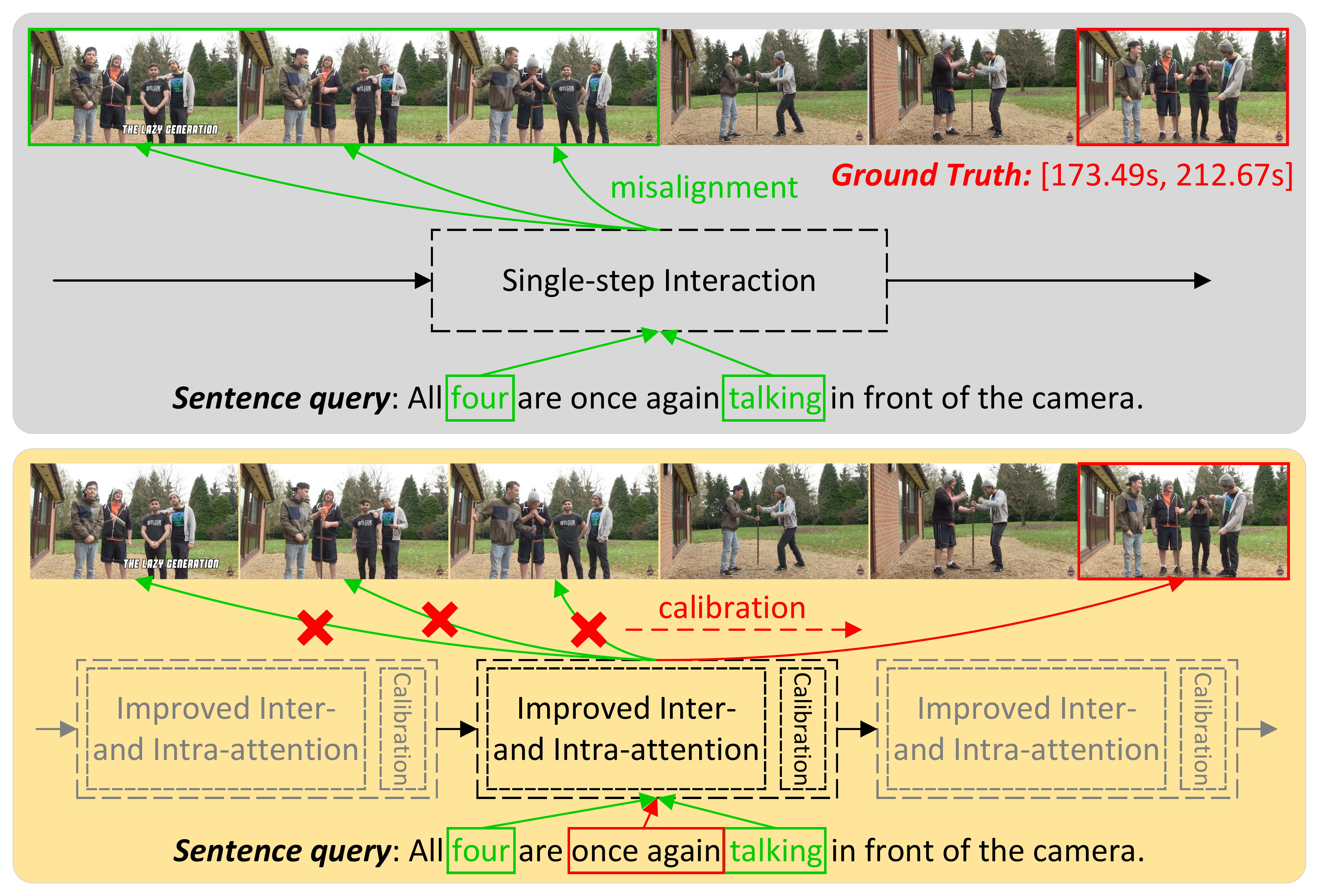}
\caption{Illustration of our motivation. \textbf{Upper}: Previous methods are mainly based on a single-step interaction with attention, which is insufficient to reason the complicated multi-modal relations, thus may lead to the misalignment on semantics. \textbf{Bottom}: we develop an iterative network with improved attention mechanism and calibration module, which can progressively align accurate semantic.}
\label{fig:introduction}
\vspace{-12pt}
\end{figure}

Existing methods mainly focus on learning multi-modal interaction by a single-step attention mechanism. Most of approaches \cite{yuan2019find,chen2019semantic,zhang2019man,rodriguez2020proposal,chenrethinking,liu2020reasoning,liu2020jointly} utilize a simple co-attention mechanism to learn the inter-modality relations between word-frame pairs for aligning the semantic information. Besides, some approaches \cite{chen2019localizing,zhang2019cross,liu2021context} employ a single-step self-attention to explore the contextual information among intra-modality to correlate relevant frames or words.

Although these methods achieve promising results, they are severely limited by two issues. 1) These single-step methods only consider the inter- or intra-modal relation once, which is insufficient to learn the complicated multi-modal interaction that needs multi-step reasoning. Besides, the misalignment between inter-modality or the wrong-attention across intra-modality caused by such single-step attention will directly degenerate the performance on the boundary results.
As shown in Figure \ref{fig:introduction}, the task targets to localize the query “\textit{All four are \textbf{once again} talking in front of the camera}” in the video. It is hard for single-step methods to directly pay more attention on phrase ”\textit{\textbf{once again}}”, easily leading to the misalignment problem and the wrong grounding result. 
2) Nowhere-to-attend problem is generally happened in TSG task, in which the background frames do not match any word in the sentence, and the basic attention may generate the wrong attention weights in these cases.

In this paper, we develop a novel Iterative Alignment Network (IA-Net) for temporal sentence grounding, which addresses the above problems by an end-to-end framework within multi-step reasoning. 
Specifically, we introduce an iterative matching scheme to explore both inter- and intra-modal relations progressively with an improved attention based inter- and intra-modal interaction module. 
In this module,
we first pad the multi-modal features with learnable parameters to tackle the nowhere-to-attend problem, and enhance the basic co-attention mechanism into a parallel manner that can provide multiple attended features for better capturing the complicated inter- and intra-modal relations. 
Then, to refine and calibrate the misaligned attention happened in early reasoning step, we develop a calibration module following each attention module to refine the alignment knowledge during the iterative process.
By stacking multiple such improved interaction modules, our IA-Net provides effective attention to iteratively reason the complicated relations between the vision and language features step-by-step, providing more accurate segment boundaries.

Our main contributions are three-fold:
\begin{itemize}
    \item We propose an iterative framework for temporal sentence grounding to progressively align the complicated semantics between vision and language.
    \item We formulate the proposed iterative matching method with an improved co-attention mechanism to utilize learable paddings to address nowhere-to-attend problem with deep latent clues, and a calibration module to refine or calibrate the alignment knowledge of inter- and intra-modal relations during the reasoning process.
    \item Extensive experiments are performed to examine the effectiveness of the proposed IA-Net on three datasets (ActivityNet Captions, TACoS, and Charades-STA), in which we achieve the state-of- the-art performances.
\end{itemize}

\section{Related Work}
\noindent \textbf{Temporal sentence grounding.}
Temporal sentence grounding (TSG) is a new task introduced recently \cite{gao2017tall,anne2017localizing}. Formally, given an untrimmed video and a natural sentence query, the task aims to identify the start and end timestamps of one specific video segment, which contains activities of interest semantically corresponding to the given sentence query. To interact video and sentence features, some works align the semantics of video with language by a recurrent neural network (RNN). \cite{chen2018temporally} design a recurrent module to temporally capture the evolving fine-grained frame-by-word interactions between video and sentence. \cite{zhang2019exploiting} propose to apply a bidirectional GRU instead of normal RNN for alignment. However, these RNNs can not align the semantics well in this task. As attention has proved its effectiveness on contextual correlation mining, amount of works tend to align relevant visual features with the query text description by an attention module. \cite{liu2018attentive} design a memory attention mechanism on query sentence to emphasize the visual features mentioned in the sentence. \cite{wang2019temporally} use a soft attention on moment features based on the sentence feature. \cite{zhang2019learning} adopt a simple multiplication operation for visual and language feature fusion. Moreover, the visual-textual co-attention module is widely utilized to model the cross-modal interaction \cite{liu2018cross,yuan2019find,chen2019localizing,chenrethinking,rodriguez2020proposal,jiang2019cross,qu2020fine,nan2021interventional}, which performs effective and efficient in most of challenging scenes. There are also some works \cite{zhang2019cross,chen2019localizing} adopt self-attention block to correlate frames or words
in each modality for constructing scene meaning.

\begin{figure*}[t!]
\centering
\includegraphics[width=1.0\textwidth]{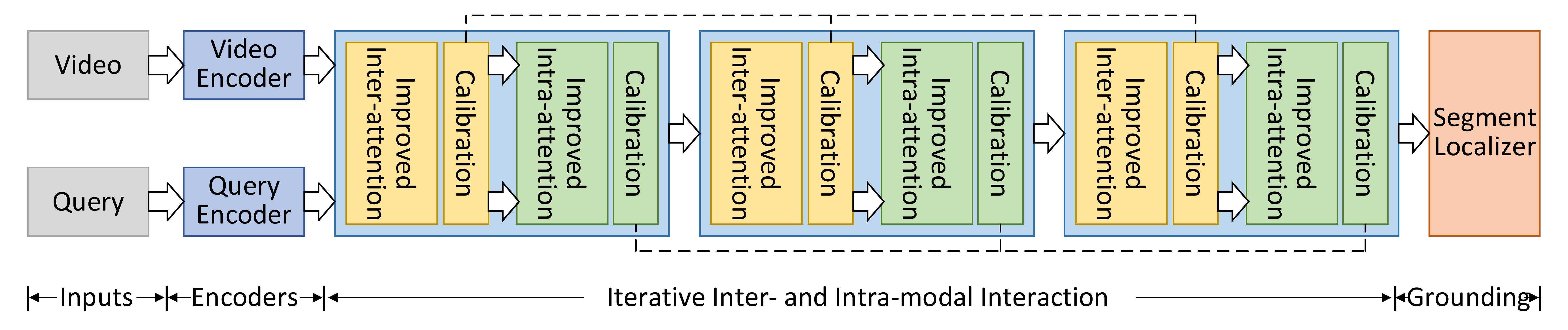}
\caption{The architecture of our proposed IA-Net. The iterative process is devised for multi-modal interaction.}
\label{fig:pipeline}
\end{figure*}

Although these attention based methods have made great progress in TSG, they are severely limited by such single-step attention mechanism. 
Motivated by this, we introduce an iterative alignment scheme to explore fine-grained inter- and intra-modal relations. For complicated correlation capturing, we pad the multi-modal features with learable parameters and enhance the co-attention with multi heads. For semantic misalignment, we additionally develop a calibration module to refine the alignment knowledge. Such iterative process helps our model align more accurate semantic.

\noindent \textbf{Attention mechanism.}
Attention has achieved great success
in various tasks, such as image classification, machine translation, and visual question answering. \cite{vaswani2017attention} propose the Transformer to capture the long-term dependency with multi-headed architecture. Although it attracts great interests from multi-modal retrieval community, it only consider sentence-guided attention on video frames or the video-guided attention on sentence words with complex computation. Compared to it, co-attention mechanism \cite{lu2016hierarchical,xiong2016dynamic} is proposed to jointly reason about frames and words attention with light weights, which is more suitable for addressing the real-world temporal sentence grounding task. 
In this paper, we consider the nowhere-to-attend cases in TSG task that frame/word is irrelevant to the whole sentence/video, and address it by utilizing a learnable paddings during the process of attention map generation. We also improve the basic co-attention mechanism into a parallel manner like Transformer which provides multiple latent attended features for better correlation mining.

\section{The Proposed Method}
The TSG task considered in this paper is defined as follows. Given an untrimmed reference video $\mathcal{V}$ and a sentence query $\mathcal{Q}$, we need to predict the start and end timestamps $(\tau_s, \tau_e)$, where the segment in $\mathcal{V}$ from time point $\tau_s$ to $\tau_e$ corresponds to the same semantic as $\mathcal{Q}$.

In this section, we introduce our framework IA-Net as shown in Figure \ref{fig:pipeline}. Our model consists of three main components: video and query encoders, iterative inter- and intra-modal interaction, and the segment localizer.
Video and sentence are first fed to the encoders for extracting multi-modal features. Then we iteratively interact their features for semantic alignment. 
Specially, in each iterative step,
we utilize a co-attention to align the inter-modal semantic and an another co-attention to correlate intra-modal instances in each modality. We improve the basic co-attention mechanism in a parallel manner, and devise two calibration modules following the inter- and intra-attention to refine and calibrate the knowledge of cross-modal alignment and self-modal correlation during the iterative interaction process. At last, we utilize a segment localizer to ground the segment boundaries.

\subsection{Video and Query Encoders}
\noindent \textbf{Video encoder.} For video encoding, we first extract the clip-wise features by a pre-trained C3D network \cite{tran2015learning}, and then add a positional encoding \cite{vaswani2017attention} to take positional knowledge. Considering the sequential characteristic in video, a bi-directional GRU \cite{chung2014empirical} is further utilized to incorporate the contextual information in time series. The output of this encoder
is $\bm{V}=\{\bm{v}_t\}^{T}_{t=1} \in \mathbb{R}^{T \times D}$, which encodes the context in video.

\noindent \textbf{Query encoder.} For query encoding, we first extract the word embeddings by the Glove word2vec model \cite{pennington2014glove}, and also use the positional encoding and bi-directional GRU to integrate the sequential information. The final feature representation of the input sentence is denoted as $\bm{Q} = \{\bm{q}_n\}_{n=1}^N \in \mathbb{R}^{N \times D}$. 

\subsection{Improved Inter-modal Interaction}
\begin{figure}[t!]
\centering
\includegraphics[width=0.48\textwidth]{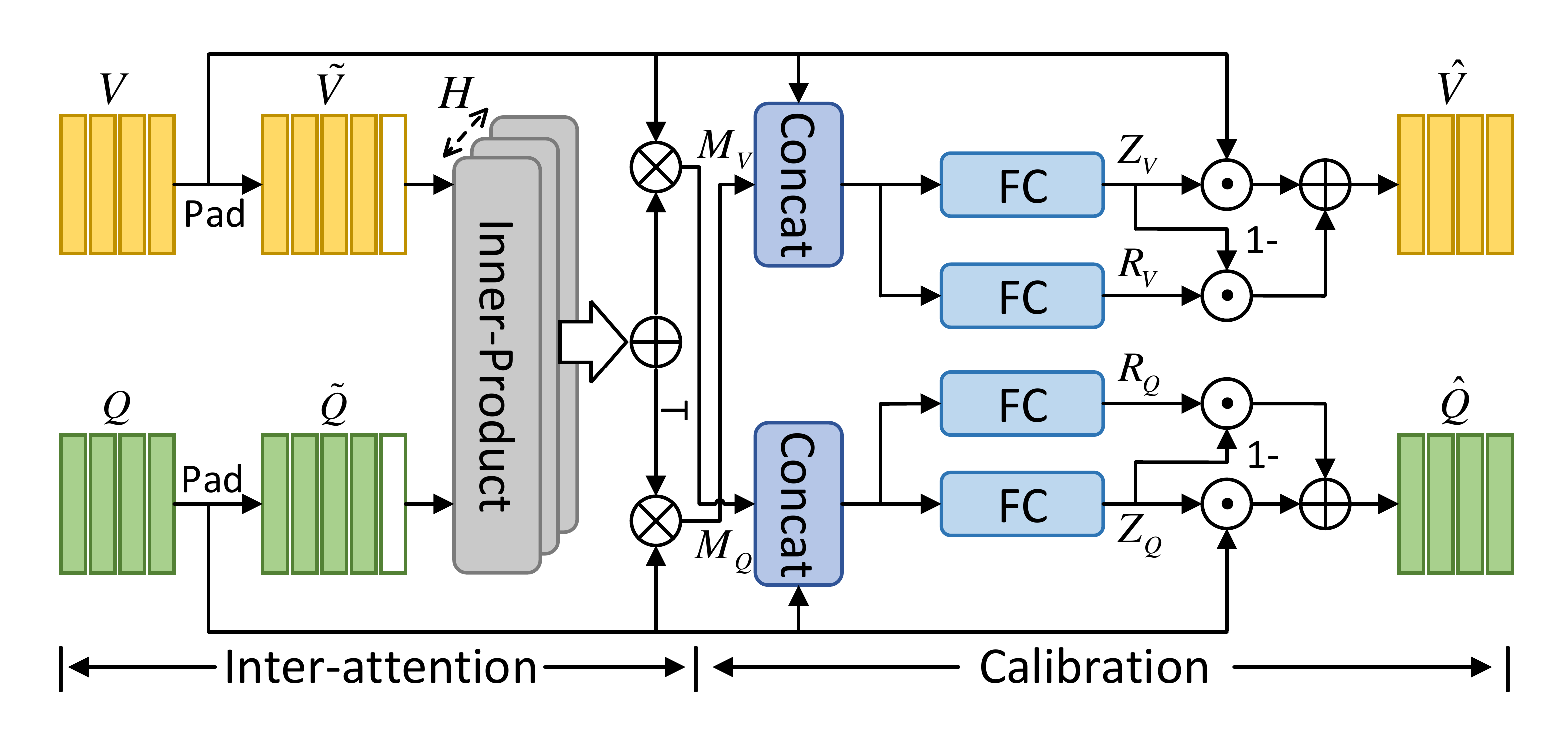}
\caption{Illustration of our inter-modal interaction.}
\label{fig:cross}
\end{figure}

The improved inter-modal interaction module is based on co-attention mechanism to capture the importance between each pair of visual clip and word features. To tackle nowhere-to-attend problem and calibrate the misalignment knowledge, we improve the co-attention in a parallel manner with learnable paddings and devise a calibration module followed by it. Details are shown in Figure \ref{fig:cross}.

\noindent \textbf{Nowhere-to-attend and parallel attention.}
Previous co-attention based works in TSG \cite{yuan2019find,chen2019localizing,chenrethinking} formally compute the attention maps by directly calculating the inner product between $\bm{V}, \bm{Q}$. However,
it often occurs at the creation of an attention map that there is no particular frame or word that the model should attend, especially for the background frames that do not match any word in the sentence. This will lead to the wrong attention on the mismatched frame-word pairs.
To deal with such cases, we add $K$ elements to both sentence words and video clips to additionally serve for no-attention instances. In details, we incorporate two learnable matrices $[\bm{v}_{{\oslash}_1},...,\bm{v}_{{\oslash}_k}], [\bm{q}_{{\oslash}_1},...,\bm{q}_{{\oslash}_k}] \in \mathbb{R}^{K \times D}$ to $\bm{V},\bm{Q}$ as $\widetilde{\bm{V}}=[\bm{v}_1,...,\bm{v}_T,\bm{v}_{{\oslash}_1},...,\bm{v}_{{\oslash}_k}] \in \mathbb{R}^{(T+K) \times D} ,\widetilde{\bm{Q}}=[\bm{q}_1,...,\bm{q}_N,\bm{q}_{{\oslash}_1},...,\bm{q}_{{\oslash}_k}] \in \mathbb{R}^{(N+K) \times D}$, respectively.
Besides, we also enhance co-attention into multiple attention heads to capture complicated relations in different latent space, and use their average as the attention result.

To generate $H$ number of attention maps, we first linearly project the $D$-dimensional features of $\widetilde{\bm{V}}, \widetilde{\bm{Q}}$ into multiple lower
$D^{-}$-dimensional spaces, where $D^{-}=D/H$. We take the $h$-th attention map ($h\in H$) as example:
\begin{equation}
    \widetilde{\bm{V}}^h = \text{Linear}(\widetilde{\bm{V}};\Theta_{V,h}), \
    \widetilde{\bm{Q}}^h = \text{Linear}(\widetilde{\bm{Q}};\Theta_{Q,h}),
\end{equation}
where $\text{Linear}(\cdot)$ denotes a fully-connected layer with parameter $\Theta$. Then, we compute the attention map by inner product with row-wise normalization as:
\begin{equation}
    \bm{A}^h_V = \text{Softmax}(\frac{\widetilde{\bm{V}}^h(\widetilde{\bm{Q}}^h)^T}{\sqrt{D^{-}}}) \in \mathbb{R}^{(T+K)\times(N+K)},
\end{equation}
\begin{equation}
    \bm{A}^h_Q = \text{Softmax}(\frac{\widetilde{\bm{Q}}^h(\widetilde{\bm{V}}^h)^T}{\sqrt{D^{-}}}) \in \mathbb{R}^{(N+K)\times(T+K)}.
\end{equation}
We take average fusion of multiple attended features, which is equivalent to averaging $H$ number of attention maps as:
\begin{equation}
    \bm{A}_V = \frac{1}{H}\sum_{h=1}^H \bm{A}^h_V, \
    \bm{A}_Q = \frac{1}{H}\sum_{h=1}^H \bm{A}^h_Q.
\label{maps}
\end{equation}
At last, we can get the $\bm{V,Q}$-grounded alignment features $\bm{M}_V, \bm{M}_Q$, in which each element captures related semantics shared by the whole $\bm{Q}, \bm{V}$ to each $\bm{v}_t,\bm{q}_n$:
\begin{equation}
    \bm{M}_V = (\bm{A}_V \widetilde{\bm{Q}})[1:T,:] \in \mathbb{R}^{T \times D},
\end{equation}
\begin{equation}
    \bm{M}_Q = (\bm{A}_Q \widetilde{\bm{V}})[1:N,:] \in \mathbb{R}^{N \times D}.
\end{equation}

\noindent \textbf{Calibration module.}
After receiving the alignment features $\bm{M}_V,\bm{M}_Q$ and the multi-modal features $\bm{V},\bm{Q}$, to refine the alignment knowledge for the next interaction step, we aim to update each modal features $\bm{V},\bm{Q}$ by aggregating them with the corresponding alignment features $\bm{M}_V,\bm{M}_Q$ with a gate function dynamically. In details, we first generate a fusion feature for each modality to enhance its semantics by:
\begin{equation}
    \bm{R}_V = \text{Tanh}(\bm{W}_{R_V}\bm{V}+\bm{U}_{R_V}\bm{M}_V+\bm{b}_{R_V}),
\end{equation}
\begin{equation}
    \bm{R}_Q = \text{Tanh}(\bm{W}_{R_Q}\bm{Q}+\bm{U}_{R_Q}\bm{M}_Q+\bm{b}_{R_Q}),
\end{equation}
where $\bm{W},\bm{U},\bm{b}$ are the learnable parameters. To select the discriminative information and filter out incorrect one, a gating weight can be formulated as follows:
\begin{equation}
    \bm{Z}_V = \text{Sigmoid}(\bm{W}_{Z_V}\bm{V}+\bm{U}_{Z_V}\bm{M}_V+\bm{b}_{Z_V}),
\end{equation}
\begin{equation}
    \bm{Z}_Q = \text{Sigmoid}(\bm{W}_{Z_Q}\bm{Q}+\bm{U}_{Z_Q}\bm{M}_Q+\bm{b}_{Z_Q}).
\end{equation}
At last, the calibrated output of the current inter-modal interaction module can be obtained by:
\begin{equation}
    \widehat{\bm{V}} = \bm{Z}_V \odot \bm{V} + (1-\bm{Z}_V) \odot \bm{R}_V,
\end{equation}
\begin{equation}
    \widehat{\bm{Q}} = \bm{Z}_Q \odot \bm{Q} + (1-\bm{Z}_Q) \odot \bm{R}_Q,
\end{equation}
where $\odot$ denotes the element-wise multiplication.

The developed gate mechanism has two main contributions: 1) The information of each modality can be refined by itself and the enhanced semantic features shared with $\bm{R}$. It helps to filter out trivial information maintained in $\bm{V},\bm{Q}$, and calibrate the misaligned attention by re-considering its individual shared semantics. 2) The contextual information from alignment features $\bm{M}_V, \bm{M}_Q$ summarize the contexts regard to each instances in cross-modal features $\bm{Q}, \bm{V}$, respectively. After the gating process, the contextual information maintained in $\widehat{\bm{V}}, \widehat{\bm{Q}}$ will also assist to determine the shared semantics in latter attention procedure. It will progressively enhance the interaction among inter-modal features and thus benefit the representation learning.

\subsection{Improved Intra-modal Interaction}
\begin{figure}[t!]
\centering
\includegraphics[width=0.48\textwidth]{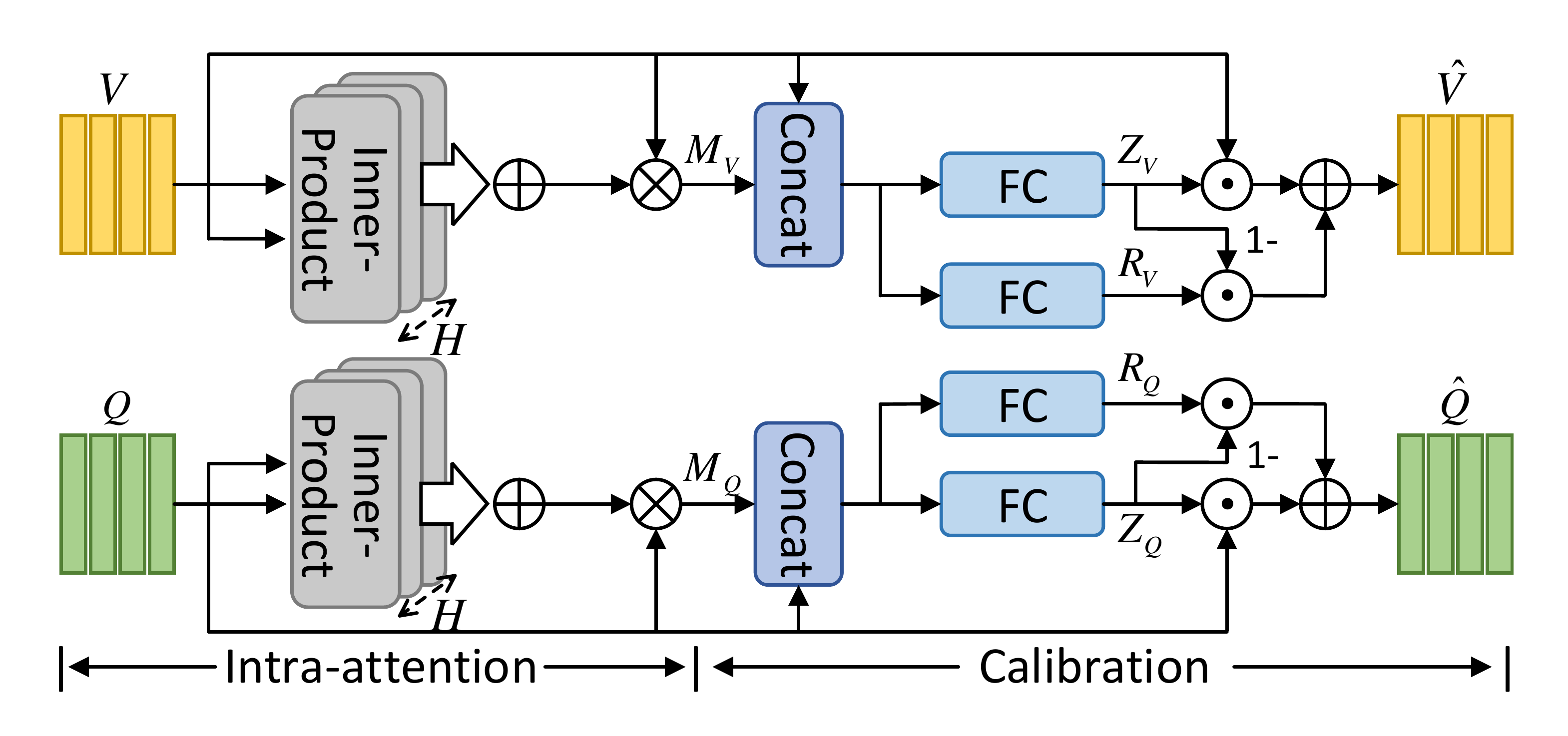}
\caption{Illustration of our intra-modal interaction.}
\label{fig:self}
\end{figure}

The output visual clip and sentence word features of the inter-modal interaction module have encoded cross-modal relations between clips and words. With such contextual cross-modal information, we implement intra-attention on each modality to correlate the relevant instances for composing the scene meaning. Different from inter-attention, there is no nowhere-to-attend problem as video or sentence has strong temporal relations in itself.
A calibration module is also utilized for self-relation refinement as shown in Figure \ref{fig:self}.

\subsection{Iterative Alignment with The Improved Inter- and Intra-modal Interaction Block}
In this section, we introduce how to integrate the improved inter- and intra-modal interaction modules to enable the iterative alignment for temporal sentence grounding.
The inter-modal interaction helps aggregate features from the other modality to update the clip and word features according to the cross-modal relations. The clip and word features would be updated again with the information within the same modality via the intra-modal interaction. We use one inter-modal interaction module followed by one intra-modal interaction module to form a basic improved interaction block (\textbf{IIB}) in our proposed IA-Net framework as:
\begin{equation}
    \widehat{\bm{V}}_l,\widehat{\bm{Q}}_l =  \textbf{IIB}(\widehat{\bm{V}}_{l-1},\widehat{\bm{Q}}_{l-1}),
\end{equation}
where $l$ is the block number.
Multiple blocks could be further stacked thanks to the calibration module for alignment refinement, helping to reason for the accurate segment boundaries.

\subsection{Segment Localizer and Loss Function}
After multiple interaction blocks, we utilize a cosine similarity function \cite{mithun2019weakly} on $\widehat{\bm{V}},\widehat{\bm{Q}}$ to generate a new video-aware sentence representation $\widehat{\bm{Q}}'$ which has the same $T$-dimensional features like $\widehat{\bm{V}}$. We fuse two modal features as $\bm{f}_t=[\widehat{\bm{v}}_t,\widehat{\bm{q}}_t'], \bm{F}=\{\bm{f}_t\}_{t=1}^T \in \mathbb{R}^{T \times 2D}$ by concatenation.
To predict the target video segment,
similar to \cite{yuan2019semantic}, we pre-define multi-size candidate moments $\{(\hat{\tau}_s, \hat{\tau}_e)\}$ on each frame $t$, and adopt multiple full connection (FC) layers to process features $\bm{f}_t$ to produce the confidence scores $\{cs\}$ of all windows and predict corresponding temporal offsets $\{(\hat{\delta}_s, \hat{\delta}_e)\}$. The final predicted moments of time $t$ can be presented as $\{(\hat{\tau}_s+\hat{\delta}_s, \hat{\tau}_e+\hat{\delta}_e)\}$.

\begin{table*}[t!]
    \small
    \centering
    \setlength{\tabcolsep}{0.8mm}{
    \begin{tabular}{cccccccccccccc}
    \hline \hline
    \multirow{3}*{Method} & \multicolumn{6}{c}{ActivityNet Captions} & & \multicolumn{6}{c}{TACoS} \\ \cline{2-7} \cline{9-14}
    ~ & R@1, & R@1, & R@1, & R@5, & R@5, & R@5, & & R@1, & R@1, & R@1, & R@5, & R@5, & R@5, \\ 
    ~ & IoU=0.3 & IoU=0.5 & IoU=0.7 & IoU=0.3 & IoU=0.5 & IoU=0.7 & & IoU=0.1 & IoU=0.3 & IoU=0.5 & IoU=0.1 & IoU=0.3 & IoU=0.5 \\ \hline
    TGN & 45.51 & 28.47 & - & 57.32 & 43.33 & - & & 41.87 & 21.77 & 18.90 & 53.40 & 39.06 & 31.02\\
    CTRL & 47.43 & 29.01 & 10.34 & 75.32 & 59.17 & 37.54 & & 24.32 & 18.32 & 13.30 & 48.73 & 36.69 & 25.42 \\
    ACRN & 49.70 & 31.67 & 11.25 & 76.50 & 60.34 & 38.57 & & 24.22 & 19.52 & 14.62 & 47.42 & 34.97 & 24.88 \\
    CBP & 54.30 & 35.76 & 17.80 & 77.63 & 65.89 & 46.20 & & - & 27.31 & 24.79 & - & 43.64 & 37.40 \\
    SCDM & 54.80 & 36.75 & 19.86 & 77.29 & 64.99 & 41.53 & &  - & 26.11 & 21.17 & - & 40.16 & 32.18\\
    ABLR & 55.67 & 36.79 & - & - & - & - & & 34.70 & 19.50 & 9.40 & - & - & - \\
    GDP & 56.17 & 39.27 & - & - & - & - & & 39.68 & 24.14 & 13.50 & - & - & - \\
    CMIN & 63.61 & 43.40 & 23.88 & 80.54 & 67.95 & 50.73 & & 32.48 & 24.64 & 18.05 & 62.13 & 38.46 & 27.02\\ 
    2DTAN & 59.45 & 44.51 & 26.54 & 85.53 & 77.13 & 61.96 & & \textbf{47.59} & 37.29 & 25.32 & 70.31 & \textbf{57.81} & 45.04 \\
    DRN & - & 45.45 & 24.36 & - & 77.97 & 50.30 & & - & - & 23.17 & - & - & 33.36 \\ \hline
    \textbf{IA-Net} & \textbf{67.14} & \textbf{48.57} & \textbf{27.95} & \textbf{87.21} & \textbf{78.99} & \textbf{63.12} & & 47.18 & \textbf{37.91} & \textbf{26.27} & \textbf{71.75} & 57.62 & \textbf{46.39}  \\ \hline
    \end{tabular}}
    \caption{Performance compared with the state-of-the-art TSG models on ActivityNet Captions and TACoS dataset.}
    \label{tab:compare1}
\end{table*}

\noindent \textbf{Training.}
We first compute the Intersection over Union score $o$ between each candidate moment $(\hat{\tau}_s, \hat{\tau}_e)$ with ground truth $(\tau_s, \tau_e)$. If the $o$ is larger than a threshold value $\lambda$, this moment is viewed as positive sample, reverse as the negative sample. Thus we can obtain $N_{pos}$ positive samples and $N_{neg}$ negative samples in total ($N_{total}$).
We adopt an alignment loss to align the predicted confidence scores and IoU:
\begin{equation}
    \mathcal{L}_{align} = -\frac{1}{N_{total}}\sum o\text{log}(cs)+(1-o)\text{log}(1-cs).
\end{equation}
We also devise a boundary loss for $N_{pos}$ positive samples to promote exploring the precise start and end points as:
\begin{equation}
    \mathcal{L}_b = \frac{1}{N_{pos}}\sum \mathcal{S}(\hat{\delta}_s - \delta_s) + \mathcal{S}(\hat{\delta}_e - \delta_e),
\end{equation}
where $\mathcal{S}$ represents the smooth L1 function. We adopt $\alpha$ to control the balance of the alignment loss and boundary loss:
\begin{equation}
    \mathcal{L} = \mathcal{L}_{align} + \alpha \mathcal{L}_b.
\end{equation}

\noindent \textbf{Testing.}
We rank all candidate moments according to their predicted confidence scores, and then “Top-n (Rank@n)” candidates will be selected with non maximum suppression.

\section{Experiments}
\subsection{Datasets}
\noindent \textbf{ActivityNet Captions.}
ActivityNet Captions \cite{krishna2017dense} contains 20k untrimmed videos with 100k descriptions from YouTube. The videos are 2 minutes on average, and the annotated video clips have much larger variation, ranging from several seconds to over 3 minutes. Following public split, we use 37,417, 17,505, and 17,031 sentence-video pairs for training, validation, and testing respectively.

\noindent \textbf{TACoS.}
TACoS \cite{regneri2013grounding} is widely used on TSG task and contain 127 videos. The videos from TACoS are collected from cooking scenarios, thus lacking the diversity. They are around 7 minutes on average. We use the same split as \cite{gao2017tall}, which includes 10146, 4589, 4083 query-segment pairs for training, validation and testing.

\noindent \textbf{Charades-STA.}
Charades-STA is built on the Charades dataset \cite{sigurdsson2016hollywood}, which focuses on indoor activities. In total, there are 12408 and 3720 moment-query pairs in the training and testing sets respectively.

\subsection{Experimental Settings}
\noindent \textbf{Evaluation Metric.}
Following previous works \cite{gao2017tall,yuan2019semantic,zhang2019learning}, we adopt “R@n, IoU=m” as our evaluation metrics. The “R@n, IoU=m” is defined as the percentage of at least one of top-n selected moments having IoU larger than m.

\noindent \textbf{Implementation Details.}
We define continuous 16 frames as a clip and each clip overlaps 8 frames with adjacent clips, and apply C3D \cite{tran2015learning} to encode the videos on ActivityNet Captions, TACoS, and I3D \cite{carreira2017quo} on Charades-STA. 
We set the length of video feature sequences to 200 for ActivityNet Captions and TACoS datasets, 64 for Charades-STA dataset. As for sentence encoding, we utilize Glove word2vec \cite{pennington2014glove} to embed each word to 300 dimension features. The hidden state dimension of Bi-GRU networks is set to 512. During segment localization, we adopt convolution kernel size of [16, 32, 64, 96, 128, 160, 192] for ActivityNet Captions, [8, 16, 32, 64, 128] for TACoS, and [16, 24, 32, 40] for Charades-STA. We set the stride of them as 0.5, 0.125, 0.125, respectively. We set the high-score threshold $\lambda$ to 0.45, and the balance hyper-parameter $\alpha$ to 0.001 for ActivityNet Captions, 0.005 for TACoS and Charades-STA. We train our model with an Adam optimizer with leaning rate $8 \times 10^{-4}$, $3 \times 10^{-4}$, $4 \times 10^{-4}$ for Activity Captions, TACoS and Charades-STA, respectively.

\subsection{Comparison to State-of-the-Art Methods}
\noindent \textbf{Compared methods.} We compare our proposed model with the following baseline methods on the TSG task: TGN \cite{chen2018temporally}, CTRL \cite{gao2017tall}, ACRN \cite{liu2018attentive}, CBP \cite{wang2019temporally}, SCDM \cite{yuan2019semantic}, ABLR \cite{yuan2019find}, GDP \cite{chenrethinking}, CMIN \cite{zhang2019cross}, 2DTAN \cite{zhang2019learning}, and DRN \cite{zeng2020dense}. These methods interact multi-modal features only in a single step process.

\noindent \textbf{Analysis.}
As shown in Tables \ref{tab:compare1} and \ref{tab:compare2}, we compare our IA-Net with all above methods on three datasets. It shows that IA-Net performs among the best
in various scenarios on all three benchmark datasets across different criteria and ranks the first or the second in all cases. On ActivityNet Captions, we outperform DRN by 3.59\% and 12.82\% in the strict metrics “R@1, IoU=0.7” and “R@5, IoU=0.7”. We also brings 1.41\% and 1.16\% improvements compared to 2DTAN. On TACoS dataset, the cooking activities take place in the same kitchen scene with some slightly varied cooking objects, thus it is hard to localize such fine-grained activities. Compared to the top ranked method 2DTAN, our model still achieves the best results on “R@1, IoU=0.5” and “R@5, IoU=0.5”, which validates that IA-Net is able to localize the moment boundary more precisely. On Charades-STA, we outperform the SCDM by 6.85\%, 4.48\%, 15.35\% and 3.96\% in all metrics.
The main reasons for our proposed model outperforming the competing models lie in two folds. First, compared to methods like GDP and CMIN which utilize basic attention to interact multi-modal features, our method provides an improved attention mechanism to address ``nowhere-to-attend" problem. We also devise a distillation module to refine and calibrate the alignment knowledge. Second, previous works all adopt a single interaction process with no tolerance on the attention mistake. Thanks to the designed distillation module across the whole framework, our IA-Net can stack multiple interaction blocks to progressively reasoning for the segment boundaries in an accurate direction.

\begin{table}[t!]
    \small
    \centering
    \setlength{\tabcolsep}{1.5mm}{
    \begin{tabular}{cccccc}
    \hline \hline
    \multirow{3}*{Method} & & \multicolumn{4}{c}{Charades-STA} \\ \cline{3-6}
    ~ & & R@1, & R@1, & R@5, & R@5,  \\ 
    ~ & & IoU=0.5 & IoU=0.7 & IoU=0.5 & IoU=0.7 \\ \hline
    CTRL & & 23.63 & 8.89 & 58.92 & 29.57 \\
    ACRN & & 20.26 & 7.64 & 71.99 & 27.79\\
    CBP & & 36.80 & 18.87 & 70.94 & 50.19\\
    GDP & & 39.47 & 18.49 & - & -\\
    2DTAN & & 39.81 & 23.25 & 79.33 & 51.15\\
    SCDM & & 54.44 & 33.43 & 74.43 & 58.08\\ \hline
    \textbf{IA-Net} & & \textbf{61.29} & \textbf{37.91} & \textbf{89.78} & \textbf{62.04} \\ \hline
    \end{tabular}}
    \caption{Performance compared with the state-of-the-art TSG models on Charades-STA dataset.}
    \label{tab:compare2}
\end{table}

\subsection{Model Efficiency Comparison}
To further investigate the efficiency of our IA-Net, we conduct the comparison on TACoS dataset with other released methods. All experiments are run on one NVIDIA TITAN-XP GPU. As shown in Table \ref{tab:compare3}, “Run-Time” denotes the average time to localize one sentence in a given video, “Model Size” denotes the size of parameters. It can be observed that our IA-Net achieves the fastest run-time with the relatively smaller model size. Since CTRL and ACRN need to sample candidate segments with various sliding windows, they need a quite time-consuming matching procedure. 2DTAN adopts a convolution architecture to generate a large 2D temporal map, which contains a large number of parameters across the convolution layers. Compared to them, our IA-Net is lightweight with a few parameters of the linear layers, leading to relatively smaller model size, thus is faster than 2DTAN. 

\begin{table}[t!]
    \small
    \centering
    \setlength{\tabcolsep}{1.5mm}{
    \begin{tabular}{ccc}
    \hline \hline
    Method & Run-Time & Model Size \\ \hline
    CTRL & 2.23s & \textbf{22M} \\ 
    ACRN & 4.31s & 128M \\
    2DTAN & 0.57s & 232M \\ \hline
    \textbf{IA-Net} & \textbf{0.11s} & 68M \\ \hline
    \end{tabular}}
    \caption{Efficiency comparison run on TACoS dataset.}
    \label{tab:compare3}
\end{table}

\begin{figure*}[t!]
\centering
\includegraphics[width=1.0\textwidth]{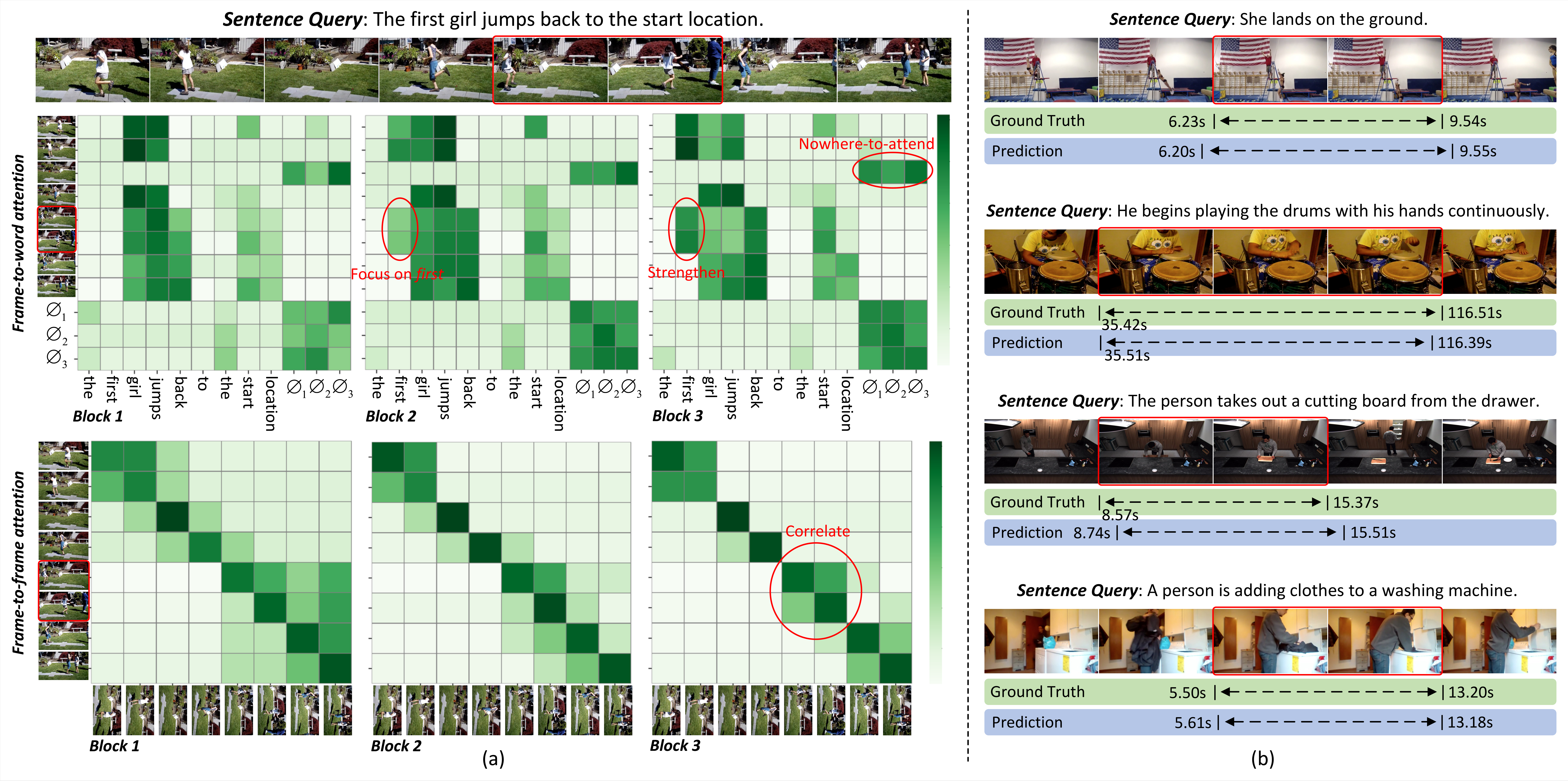}
\caption{(a) Visualization on the inter- and intra-attention of different interaction blocks. (b) Qualitative results.}
\label{fig:result}
\end{figure*}

\subsection{Ablation Study}
We perform extensive ablation studies on the ActivityNet Captions dataset. The results are shown in Table \ref{tab:ablation}.

\noindent \textbf{Effectiveness of each component.} 
To evaluate each component in our improved interaction block, we design four strong baselines: model A only contains the improved inter-attention module for multi-modal interaction; model B adds the calibration module after each inter-attention module of model A; model C adds an improved intra-attention module to model B; model D is the IA-Net, which has additional calibration module after each intra-attention module of model C. 
For fair comparison, we stack the same number of interaction blocks in all four baselines for evaluation. 
From the table, we can find that both two calibration modules in each interaction block of IA-Net can bring significant improvement (A$\to$B, C$\to$D) by refining the alignment knowledge. The intra-attention can also improve the performance (B$\to$C) by capturing the contextual information among intra-modality.

\noindent \textbf{How to choose the padding size?} We investigate the performance on different padding size $K$, which is originally introduced to deal with “nowhere-to-attend” problem. Although $K=1$ has the ability to guide the frame or words to attend nothing, such limited latent space can not meet complicated relations between different videos and sentences.
As shown in Table \ref{tab:ablation}, we found
that the use of $K>1$ improves performance to a certain extent, and $K=3$ yields the best.

\noindent \textbf{How to choose the number of attention maps?} 
Different number of parallel attention maps will guide the model attend to different relationships in several latent spaces.
Following \cite{vaswani2017attention}, we implement 1, 2, 4, 8 number of attention maps for experiments. We find that $H=4$ achieves the best performance.

\noindent \textbf{Choice of the number of stacked interaction blocks.}
Our improved interaction block contains calibration module for refining the alignment knowledge. The results in table indicates that the model with more than 3 blocks will not bring more improvement. 

\begin{table}[t!]
    \small
    \centering
    \setlength{\tabcolsep}{1.5mm}{
    \begin{tabular}{lccc}
    \hline
    \multirow{2}*{Component} & \multirow{2}*{Details} & R@1, & R@5, \\
    ~ & ~ & IoU=0.7 & IoU=0.7 \\ \hline
    Full model & A & 21.76 & 56.60 \\
    ~ & B & 23.57 & 58.33 \\
    ~ & C & 25.62 & 61.24 \\
    ~ & D & \textbf{27.95} & \textbf{63.12} \\ \hline
    Padding size $K$ & 0 & 25.81 & 59.07\\
    ~ & 1 & 27.03 & 61.61\\
    ~ & 2 & 27.69 & 63.07 \\
    ~ & 3 & \textbf{27.95} & \textbf{63.12} \\
    ~ & 4 & 27.88 & 62.94 \\ \hline
    Number ($H$) of & 1 & 25.89 & 61.13 \\
    attention maps & 2 & 27.14 & 62.39 \\
    ~ & 4 & \textbf{27.95} & \textbf{63.12}\\
    ~ & 8 & 27.66 &  63.05\\ \hline
    Number ($l$) of & 1 & 25.17 & 59.76 \\
    stacked blocks & 2 & 26.88 & 61.63\\
    ~ & 3 & 27.95 & \textbf{63.12} \\
    ~ & 4 & \textbf{27.97} & 63.03 \\ \hline
    \end{tabular}}
    \caption{Ablation study on ActivityNet Captions.}
    \label{tab:ablation}
\end{table}

\subsection{Qualitative Results}
We visualize the fused attention maps in Figure \ref{fig:result} (a). For frame-to-word attention, it can be observed that the first interaction block fail to focus on the word ``first". With the help of the calibration module, the attention is progressively calibrated on the word ``first" in the following blocks. Besides, the third frame pays more attention on the padding elements as it does not match any words in the query, which indicates that our IA-Net addresses the ``nowhere-to-attend" problem well. For frame-to-frame attention, our model correlates the relevant frames more precisely in deeper blocks. The qualitative results are shown in Figure \ref{fig:result} (b).

\section{Conclusion}
In this paper, we have studied the problem of temporal sentence grounding, and proposed a novel Iterative Alignment Network (IA-Net) in an end-to-end fashion. The core
of our network is the multi-step reasoning process with the improved inter- and intra-modal interaction module which is designed in two aspects: 1) we pad the multi-modal features with learnable parameters for capturing more complicated correlation in deep latent space; 2) we develop a calibration module to refine and calibrate the alignments knowledge from early steps. By stacking multiple such interaction modules, our IA-Net can progressively capture the fine-grained interactions between two modalities, providing more accurate video segment boundaries. Extensive experiments on three challenging benchmarks demonstrate the effectiveness and efficiency of the proposed IA-Net.

\section{Acknowledgments}

This work was supported in part by the National Natural Science Foundation of China under grant No. 61972448.

\bibliography{anthology}
\bibliographystyle{acl_natbib}

\end{document}